\documentclass[11pt]{article}

\usepackage[final]{acl}
\usepackage{booktabs} 
\usepackage{times}
\usepackage{latexsym}
\usepackage{makecell}
\usepackage{amsmath}
\usepackage{amsfonts}
\usepackage{tcolorbox}

\usepackage[T1]{fontenc}
\usepackage[utf8]{inputenc}

\usepackage{microtype}
\usepackage{multirow}

\usepackage{inconsolata}

\usepackage{graphicx}

\title{Hierarchical Wasserstein Merging for Multi-Domain Multi-Task Learning: From Specialists to a Generalist}

\author{
\bf Ming Cheng$^1$, Jiaying Gong$^2$\thanks{This work was completed prior to joining Amazon, and does not relate to the author's position at Amazon}, Hoda Eldardiry$^1$ \\
$^1$Virginia Tech, $^2$Amazon \\
\texttt{\{ming98,hdardiry\}@vt.edu, gojiayin@amazon.com}
}

\begin{document}
\maketitle
\begin{abstract}
Multi-domain multi-task learning (MD-MTL) aims to build a single generalist model that performs well across heterogeneous domains and tasks.
However, joint training often suffers from interference under distribution shifts. Existing model merging methods mostly operate on model parameters while overlooking the geometric structure of latent representation distributions across domains and tasks.
To address these limitations, we propose Hierarchical Wasserstein Merging (HWM), a representation-level framework that models each domain-task specialist as a distribution of hidden representations on a shared support.
HWM constructs task-level and global Wasserstein barycenters to capture within-task domain variation and cross-task structure, enabling either training-free specialist aggregation by Wasserstein-derived weights or training-based generalist learning through a hybrid Wasserstein alignment loss. 
Experiments on four NLP tasks across four domains per task show that HWM achieves superior effectiveness and generalization capability in MD-MTL settings.

%We propose \textbf{Hierarchical Wasserstein Merging (HWM)}, a representation-level framework that models each domain--task specialist as a distribution of hidden representations on a shared support and uses entropically regularized optimal transport for alignment. 
\end{abstract}

\section{Introduction}
Multi-domain and multi-task learning (MD-MTL) aims to improve learning by sharing knowledge across heterogeneous domains and tasks. 
Learning one unified model that can handle multiple domains and tasks offers the potential for better generalization~\cite{DBLP:journals/corr/YangH14a}.
However, jointly learning across domains and tasks remains challenging as differences in data distributions and linguistic patterns can lead to interference and negative transfer when domains or tasks are dissimilar.

The goal of MD-MTL is to integrate knowledge from a diverse set of specialists (domain and task-specific fine-tuned LLMs) into a single generalist model with broad multi-domain and multi-task capabilities.
A straightforward solution is data merging, where the LLM is trained directly on the union of all available datasets.
While simple, this approach requires access to large quantities of labeled data simultaneously, careful balancing of task objectives, and domain and task identifiers during inference.
%MTL methods mitigate interference through architectural routing or optimization-based 
Recent work on model merging tends to integrate multiple specialist models into a unified model without extensive retraining.
Existing merging techniques typically operate at the parameter level, task level, layer level, or neuron level, while overlooking how knowledge is encoded in the model's representation distributions. 
Directly interpolating or aligning parameters may fail to preserve the geometric structure of latent representations, particularly when specialists are trained on heterogeneous domains and tasks.
%To address the above limitations, we introduce Hierarchical Wasserstein Merging (HWM), a representation-level framework that composes multi-domain multi-task specialists through distributional alignment.
This motivates a distributional view of MD-MTL model merging. In heterogeneous domain-task settings, the differences among specialists are not only reflected in individual model parameters, but also in how each specialist organizes inputs in its latent representation space. A single feature vector or a linear
parameter average may obscure domain-specific structures, whereas a probability distribution over representations captures the geometry and spread of specialist behavior on a shared probing support. %Modeling specialists as distributions therefore provides a natural way to compare and aggregate models while preserving domain- and task-specific representation structure.

There are three main challenges for MD-MTL merging. First is cross-domain geometry mismatch. Different domains can induce distinct latent structures, causing representation distributions to shift. The second challenge is cross-task hierarchy, where tasks may share partial structure (i.e., information extraction and question answering share factual grounding) while also requiring task-specific semantics. The third challenge is that a generalist is expected to integrate all specialists without requiring domain or task identifiers at inference time.

%To address these challenges, we propose Hierarchical Wasserstein Merging (HWM), a representation-level framework for MD-MTL model merging through distributional alignment. Instead of assuming that specialists are linearly compatible in parameter space, HWM represents each domain-task specialist as a probability distribution of hidden representations on a shared support. This allows model similarity and aggregation to be defined by the geometry of latent representations rather than by direct parameter interpolation. Using the entropically regularized Wasserstein distance, HWM constructs task-level barycenters that aggregate domains within each task and a global barycenter that captures shared structure across tasks. This hierarchical construction naturally matches the MD-MTL structure, domain variation is first modeled within each task, and cross-task commonality is then modeled at the global level.

To address these challenges, we propose Hierarchical Wasserstein Merging (HWM), a representation-level framework for MD-MTL model merging through distributional alignment.
HWM characterizes each domain-task specialist by the distribution of its hidden representations on a shared support. Using the entropically regularized Wasserstein distance, HWM constructs task-level barycenters that aggregate domains within each task and a global barycenter that captures shared structure across tasks. This hierarchical construction models both within-task domain diversity and cross-task relationships in a geometry-aware manner.
By operating on representation distributions, HWM captures how each specialist organizes semantic information, enabling a more stable and interpretable aggregation mechanism.
HWM supports two integration strategies: (1) For the non-training method, Wasserstein-derived hierarchical weights are used to aggregate specialists into a generalist without additional training. (2) For the training-based method, a generalist model is trained using supervised fine-tuning augmented with a Wasserstein alignment loss to align with precomputed task-level and global barycenters.
Both strategies operate without requiring domain or task identifiers during inference.
%Extensive experiments across multiple tasks and domains demonstrate that HWM outperforms existing model merging baselines and yields strong generalization in multi-domain multi-task settings.
%The results suggest that representation-level, geometry-aware aggregation provides a principled and scalable alternative to parameter-centric model merging for MD-MTL.
The main contributions of this paper are summarized as:
\begin{itemize}
    \item We propose Hierarchical Wasserstein Merging (HWM), a representation-level framework for multi-domain multi-task model merging based on distributional alignment.
    \item We develop non-training-based and training-based variants of HWM to support both specialists aggregation via Wasserstein-derived weights and generalist learning through a hybrid supervised and alignment loss. %, without requiring domain or task identifiers during inference.
    \item Extensive experiments across multiple tasks and domains demonstrate the superior effectiveness and generalization capability of HWM in multi-domain multi-task settings.
\end{itemize}

%ur goal is to (1) identify which domain-task specialists are most relevant to the target domain by comparing their induced representation distributions, and (2) construct a target representation that aggregates the selected expertise.
%This target representation may then be used either to guide the weighting of specialists or to serve as a distillation target for training a single generalist model.

%This scheduling stabilizes training by allowing the model to first learn task-relevant supervision before enforcing distributional alignment.

\section{Related Work}
\subsection{Multi-Domain Multi-Task Learning}
Multi-Domain Learning (MDL) aims to learn representations that generalize across domains through domain adaptation~\cite{ding-etal-2025-3ds, zhang-etal-2025-advancing-smoe, sun-etal-2025-adversarial, reheman-etal-2025-enhancing} and domain generalization~\cite{yu2025learning, liu-etal-2025-scaling-temporal, zhao-etal-2025-seeking, cheng2024disentangled}.
Multi-Task Learning (MTL) jointly optimizes related tasks to capture complementary knowledge~\cite{chen2024multi}.
Existing MTL methods are typically architecture-based~\cite{pmlr-v267-kong25a, xu2025multi, qi-etal-2025-generalizable, calvo-etal-2025-beyond, zhu2024uni, liu2024moe, xu2024mome} and optimization-based methods~\cite{zhang-etal-2025-mazo, li2025competitive, hu-etal-2025-impartial, gong-etal-2024-coba}. Recent studies combine MDL and MTL to exploit cross-domain and cross-task synergies~\cite{zhang2024m3oe, kwan2024m4le, xin2024mmap}. Inspired by recent advances in model merging, we introduce hierarchical Wasserstein merging, a geometry-aware framework for composing multi-domain multi-task specialists into a unified generalist model.

\subsection{Model and Representation Merging}
To integrate specialized knowledge from multiple sources, a data-centric strategy of data merging~\cite{jiang-etal-2025-developing, corrado-etal-2025-automixalign} trains a unified generalist model on the union of heterogeneous datasets.
To enable the training-free integration of fine-tuned models, model merging~\cite{sun2025towards, fu-etal-2025-training, kirsch-etal-2025-pm3, zhang-etal-2025-dynamic-task, xie-etal-2025-bone, qiao-etal-2025-seqmmr} and LoRA merging~\cite{ledoyen-etal-2025-facilitating, cheng-etal-2025-sci, zhao2025mosld, zhao2025each, liu-etal-2025-r, zhang-etal-2025-mixture, zhang-zhou-2025-unraveling} combine the capabilities of multiple fine-tuned models or adapters into a single model without additional task-specific training~\cite{fu-etal-2025-training}.

However, existing merging techniques primarily operate at parameter level~\cite{li-etal-2025-frobenius, qu-etal-2025-uq, yu2024language, jang2024model, yadav2023ties, wortsman2022model}, task level~\cite{qiu-etal-2025-superpose, akiba2025evolutionary, yu2024language, davari2024model}, layer level~\cite{wang-etal-2025-recall, yang2024adamerging}, or even neuron level~\cite{fang-etal-2025-see}.
These approaches interpolate or align model weights while overlooking how knowledge is encoded in hierarchically organized representation distributions.
Inspired by optimal transport~\cite{imfeld2024transformer, singh2020model}, Hierarchical Wasserstein Merging merges models in representation distributions over specialist embeddings, enabling a generalist model to preserve hierarchical geometric consistency across domains and tasks.

\section{Methodology}
\begin{figure*}[t!] 
 \center{\includegraphics[height=8cm,width=\textwidth]{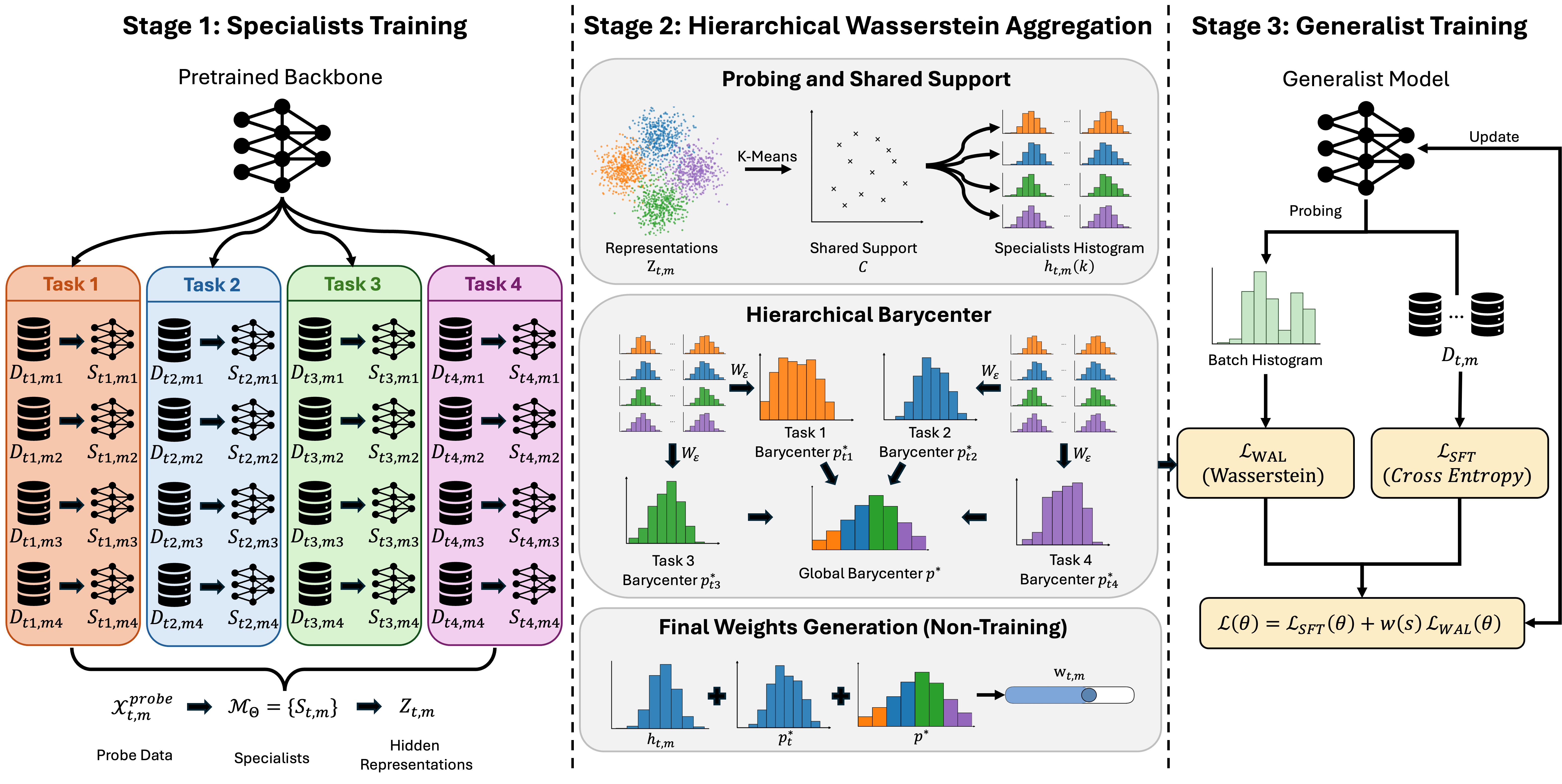}}
 \caption{\label{fig:framework} Overview of the proposed \textbf{Hierarchical Wasserstein Merging (HWM)} framework.
\textbf{Stage 1:} Domain--task specialists ($S_{t,m}$) are fine-tuned from a shared pretrained backbone.
\textbf{Stage 2:} Hidden representations are probed and aggregated on a shared support via entropic Wasserstein barycenters, resulting in task-level barycenters ($p_t^\ast$) and a global barycenter ($p^\ast$).
\textbf{Stage 3:} A single generalist model is trained using a supervised loss ($\mathcal{L}_{\mathrm{SFT}}$) augmented with a Wasserstein alignment loss ($\mathcal{L}_{\mathrm{WAL}}$), which aligns the batch-level representation distribution ($h_\theta$) to the precomputed task-level and global structures.}
 \end{figure*}

\subsection{Preliminaries}
\subsubsection{Wasserstein Distance and Barycenters}
Wasserstein barycenters extend the classical notion of averaging from individual points to probability distributions or point clouds~\cite{altschuler2021wasserstein}. 
Given a set of probability distributions $\{\mu_i\}_{i=1}^n$, the Wasserstein barycenter is defined as the distribution $\nu$ that minimizes:
\begin{equation}
\nu
=
\arg\min_{\nu}
\sum_{i=1}^{n}
\lambda_i \, \mathcal{W}(\mu_i, \nu)
\end{equation}
where $\{\lambda_i\}_{i=1}^n$ denotes non-negative weight satisfying $\sum_i \lambda_i = 1$, $\mathcal{W}(\cdot, \cdot)$ is the Wasserstein distance between distributions of hidden representations~\cite{agueh2011barycenters}. The 2-Wasserstein distance can be defined as:
\begin{equation}
W_2^2(\mu, \nu)
=
\inf_{\gamma \in \Pi(\mu, \nu)}
\mathbb{E}_{(x,y)\sim \gamma}
\left[ \lVert x - y \rVert_2^2 \right]
\end{equation}
where $\Pi(\mu, \nu)$ denotes the set of all joint probability measures with marginals $\mu$ and $\nu$, $\lVert \cdot \rVert_2$ is the $\ell_2$ norm, and $\mathrm{E}_{(x,y)\sim \gamma}[\cdot]$ denotes the expectation taken with respect to $\gamma$. In this paper, we compute a regularized approximation $\lim_{\varepsilon \to 0} W_{\varepsilon}(p, q) = W_2(p, q)$ for efficiency and stability.

%Wasserstein barycenters provide a natural extension of the notion of averaging points to the notion of averaging point clouds. 
%Importantly, they naturally inherit the ability of optimal transportation to capture geometric properties of the data.

\subsubsection{Problem Formulation}
Multi-domain multi-task learning (MD-MTL) aims to learn a single generalist model by integrating knowledge from multiple domains and task-specific specialists, where domains exhibit diverse data distributions and tasks differ in learning objectives. 
Formally, we consider a set of tasks 
$\mathcal{T} = \{\mathcal{T}_1, \mathcal{T}_2, \ldots, \mathcal{T}_T\}$ 
and a set of domains 
$\mathcal{D} = \{\mathcal{D}_1, \mathcal{D}_2, \ldots, \mathcal{D}_M\}$. 
Each task--domain pair $(\mathcal{T}_t, \mathcal{D}_m)$ is associated with a dataset
$
\mathcal{D}_{t,m} = \{(x_i^{t,m}, y_i^{t,m})\}_{i=1}^{N_{t,m}},
$
where $x_i^{t,m}$ denotes the input and $y_i^{t,m}$ denotes the corresponding output.
The goal of MD-MTL is to learn a single generalist model $F_\theta$ that simultaneously satisfies the requirements across all domains and tasks.

\subsection{Hierarchical Wasserstein Merging}
In this section, we describe our proposed HWM as shown in Figure~\ref{fig:framework} for models merging in multi-domain and multi-task scenarios. Our proposed method HWM contains three major stages: the specialists' fine-tuning stage (Sec.~\ref{sec:specialist}), the representation-level hierarchical Wasserstein aggregation stage (Sec.~\ref{sec:aggregation}), and the single generalist training stage (Sec.~\ref{sec:generalist})

\subsubsection{Specialists Training}~\label{sec:specialist}
The proposed model consists of $n$ domain-task specialists, where we primarily consider $n = 16$ in this paper across 4 tasks and 4 domains per task.
The specialists are collectively denoted as:
\begin{equation}
    \mathcal{M}_{\Theta} = \{ S_{t,m} \mid t \in \mathcal{T},\; m \in \mathcal{D}_t \}
\end{equation}
where $t \in \mathcal{T}$ denotes the task and $m \in \mathcal{D}_t$ denotes the domain associated with task $t$. Details of $\mathcal{T}$ and $\mathcal{D}_t$ are summarized in Table~\ref{tab:dataset} in the Appendix~\ref{sec:dataset_description}.
Each specialist $S_{t,m}$ is independently fine-tuned on its corresponding domain-task dataset $\mathcal{D}_{t,m}$.
We adopt Llama-3.1-8B-Instruct as the pretrained backbone for all $S_{t,m}$ due to its strong instruction-following and generalization ability~\cite{dubey2024llama}.

\begin{table*}[t]
\centering
\caption{Overview of the 30 publicly available datasets used across four NLP tasks in multiple domains.}
\scriptsize
\setlength{\tabcolsep}{4pt}
\begin{tabular}{l l l}
\toprule
\textbf{Task} & \textbf{Domain} & \textbf{Datasets Used} \\
\midrule
\multirow{4}{*}{Paraphrasing} 
 & \textit{Biomedical} & eLife~\cite{goldsack-etal-2022-making}, PLOS~\cite{goldsack-etal-2022-making}, Cells~\cite{GUO2024104580} \\
 & \textit{News} & SciTechNews~\cite{cardenas-etal-2023-dont}, MSRP~\cite{dolan-brockett-2005-automatically} \\
 & \textit{Science} & VTechAGP~\cite{cheng-etal-2025-vtechagp}, ParaSci~\cite{dong-etal-2021-parasci} \\
 & \textit{Wikipedia} & PAWS~\cite{zhang-etal-2019-paws} \\
\midrule
\multirow{4}{*}{Question Answering} 
 & \textit{Web} & ComplexWebQuestions~\cite{talmor-berant-2018-web}, WebQuestions~\cite{berant-etal-2013-semantic}, TriviaQA~\cite{joshi-etal-2017-triviaqa} \\
 & \textit{Biomedical} & COVID-QA~\cite{moller-etal-2020-covid}, HealthQA~\cite{10.1145/3308558.3313699} \\
 & \textit{Wikipedia} & WikiQA~\cite{yang-etal-2015-wikiqa}, Natural Questions~\cite{kwiatkowski-etal-2019-natural} \\
 & \textit{News} & NewsQA~\cite{trischler-etal-2017-newsqa} \\
\midrule
\multirow{4}{*}{Summarization} 
 & \textit{News} & Newsroom~\cite{grusky-etal-2018-newsroom}, Multi-News~\cite{fabbri-etal-2019-multi}, WCEP-MDS~\cite{gholipour-ghalandari-etal-2020-large} \\
 & \textit{Literature} & BookSum~\cite{kryscinski-etal-2022-booksum}, SQuALITY~\cite{wang-etal-2022-squality} \\
 & \textit{Government} & BillSum~\cite{kornilova-eidelman-2019-billsum}, BigPatent~\cite{sharma-etal-2019-bigpatent}, GovReport~\cite{huang-etal-2021-efficient} \\
 & \textit{Science} & 
   \makecell[l]{ArXiv \& PubMed~\cite{cohan-etal-2018-discourse}, SciSummNet~\cite{10.1609/aaai.v33i01.33017386},\\
   BigSurvey~\cite{ijcai2022p591}, Multi-XScience~\cite{lu-etal-2020-multi-xscience}, SumPubMed~\cite{gupta-etal-2021-sumpubmed}} \\
\midrule
\multirow{4}{*}{Information Extraction} 
 & \textit{News} & IEPile (DuIE2.0, NYT, NYT11-HRL, CoNLL2004, KBP37, GIDS, CLUE, MSRA, ACE2005, CoNLL2003, DuEE1.0) \\
 & \textit{Biomedical} & IEPile (CMeIE, ADE Corpus, PHEE, AnatEM, BC2GM, BC4CHEMD, BC5CDR, NCBI-Disease, GENIA) \\
 & \textit{Science} & IEPile (SemEval RE, SciERC, FabNER, CASIE) \\
 & \textit{Wikipedia} & IEPile (MultiNERD)~\cite{gui-etal-2024-iepile} \\
\bottomrule
\end{tabular}

\label{tab:dataset}
\end{table*}

\subsubsection{Representation-Level Wasserstein Aggregation}~\label{sec:aggregation}
\paragraph{Probing Distribution and Shared Support.}
To enable representation-level comparison across specialists, we first probe their internal activations.
For each specialist \( S_{t,m} \), we collect a probe set \( \mathcal{X}_{t,m}^{\mathrm{probe}} \) and extract hidden representations from the final transformer layer, which captures high-level semantic information.
Given an input sequence $x \in \mathcal{X}_{t,m}^{\mathrm{probe}}$, let \( \phi_\ell(x) = \{h_1, \ldots, h_{L_x}\} \) denote the sequence of hidden states at layer \( \ell \). The probed representation set for each specialist is:
\begin{equation}
Z_{t,m}
=
\left\{
\frac{1}{L_x} \sum_{i=1}^{L_x} h_i
\;\middle|\;
x \in \mathcal{X}_{t,m}^{\mathrm{probe}}
\right\}
\end{equation}
where \( L_x \) is the number of tokens in \( x \).
We further do dimensionality reduction via principal component analysis to facilitate scalable clustering and optimal transport computation.
Next, we construct a shared discrete support over the representation space by applying k-means clustering to get cluster centers \( \mathcal{C} = \{c_1,\ldots,c_K\} \).
Each specialist is then represented as a probability distribution over the shared support. We compute a soft histogram $h_{t,m}(k)$ as the empirical distributional representation of specialist $S_{t,m}$ on the shared support $\mathcal{C}$:
\begin{equation}~\label{equ:kmeans}
\small
h_{t,m}(k)
=
\frac{1}{|Z_{t,m}|}
\sum_{z \in Z_{t,m}}
\frac{
\exp\!\left(-\|z - c_k\|_2^2 / (2\tau^2)\right)
}{
\sum_{j=1}^K
\exp\!\left(-\|z - c_j\|_2^2 / (2\tau^2)\right)
}
\end{equation}
where $\tau > 0$ controls the smoothness, and $k \in \{1, \cdots K\}$ indexes the cluster centroids.

\paragraph{Hierarchical Wasserstein Aggregation.}
To aggregate domain-task distributions, we compute entropic Wasserstein barycenters on the shared support $\mathcal{C}$. Given histogram representations $\{h_{t,m}\}$ and entropic kernel $K \in \mathbb{R}^{K \times K}$, where \( K_{ij} = \exp(-\|c_i - c_j\|_2^2 / \varepsilon) \), we first compute a task-level barycenter $p_t^\ast$ by aggregating all domain distributions associated with the same task $t$:
\begin{equation}
p_t^\ast
=
\arg\min_{p \in \mathcal{P}(\mathcal{C})}
\sum_{m \in \mathcal{M}_t}
W_\varepsilon(p, h_{t,m})
\end{equation}
where \( W_\varepsilon(\cdot,\cdot) \) denotes the entropic Wasserstein distance computed with kernel \( K \).
We aggregate task-level barycenters $p_t^\ast$ to obtain a global barycenter $p^\ast$ that integrates information across all tasks:
\begin{equation}
p^\ast
=
\arg\min_{p \in \mathcal{P}(\mathcal{C})}
\sum_{t \in \mathcal{T}}
W_\varepsilon(p, p_t^\ast)
\end{equation}
Both barycenters are computed using Sinkhorn fixed-point iterations on the shared support, leading to stable and scalable solutions.
For a non-training-based HWM method, the final weight $w_{t,m}$ assigned to the specialist model $S_{t,m}$ is obtained by combining task-level and within-task relevance through a hierarchical weighting method:
\begin{equation}~\label{equ:weights}
\begin{aligned}
w_{t,m}
&=
\frac{
\exp\!\big(-\beta_{\mathrm{task}}\, W_\varepsilon(p_t^\ast, p^\ast)\big)
}{
\sum_{t' \in \mathcal{T}}
\exp\!\big(-\beta_{\mathrm{task}}\, W_\varepsilon(p_{t'}^\ast, p^\ast)\big)
}
\\[4pt]
&\quad \times
\frac{
\exp\!\big(-\beta_{\mathrm{in}}\, W_\varepsilon(h_{t,m}, p_t^\ast)\big)
}{
\sum_{m' \in \mathcal{M}_t}
\exp\!\big(-\beta_{\mathrm{in}}\, W_\varepsilon(h_{t,m'}, p_t^\ast)\big)
}
\end{aligned}
\end{equation}
where $\sum_{t \in \mathcal{T}}\sum_{m \in \mathcal{M}_t} w_{t,m} = 1$, $\beta_{\mathrm{task}}$ and $\beta_{\mathrm{in}}$ control the sharpness of task-level and within-task selection, and $W_\varepsilon(\cdot,\cdot)$ denotes the entropically regularized Wasserstein distance computed on the shared support.
This final weight prioritizes specialists whose representation distributions are geometrically closer to the corresponding task-level barycenter and the global barycenter.
%the distance between each specialist $S_{t,m}$ to the corresponding task-level barycenter $p_t^\ast$ and the global barycenter $p^\ast$.

\subsubsection{Generalist Training}~\label{sec:generalist}
We train a single generalist model using standard supervised fine-tuning augmented with a representation-level Wasserstein alignment loss.
The generalist model is initialized from the same pretrained backbone as specialists $\mathcal{M}_{\Theta}$ and optimized over labeled multi-domain and multi-task data.
The overall training objective is defined as:
\begin{equation}
\mathcal{L}(\theta)=\mathcal{L}_{\mathrm{SFT}}(\theta) + w(s)\,\mathcal{L}_{\mathrm{WAL}}(\theta) 
\end{equation}
where $\theta$ denotes the parameters of the generalist model, $\mathcal{L}_{\mathrm{SFT}}(\theta)$ is the standard supervised loss, and $\mathcal{L}_{\mathrm{WAL}}(\theta)$ is the Wasserstein loss for aligning the generalist with both task-specific and global distributional structures.
$w(s)$ is a scheduling function of the optimization step $s$ that gradually increases the influence of the Wasserstein loss after an initial warm-up period, which stabilizes optimization by allowing the model to first learn task-relevant supervision before enforcing distributional alignment.
Specifically, $\mathcal{L}_{\mathrm{SFT}}(\theta)$ is the cross-entropy loss computed on labeled training data, and $\mathcal{L}_{\mathrm{WAL}}(\theta)$ is a weighted combination of task-level and global Wasserstein alignment losses:
\begin{equation}~\label{equ:training}
\mathcal{L}_{\mathrm{WAL}}(\theta)=\lambda_{\mathrm{task}} \, \mathcal{L}_{\mathrm{WAL}}^{\mathrm{task}}(\theta)+\lambda_{\mathrm{global}} \, \mathcal{L}_{\mathrm{WAL}}^{\mathrm{global}}(\theta)
\end{equation}
where $\lambda_{\mathrm{task}}$, and $\lambda_{\mathrm{global}}$ are the weighting parameters. Specifically, $\mathcal{L}_{\mathrm{WAL}}^{\mathrm{task}}(\theta) = W_\varepsilon(h_\theta, p_t^\ast)$, and $\mathcal{L}_{\mathrm{WAL}}^{\mathrm{global}}(\theta) = W_\varepsilon(h_\theta, p^\ast)$, where $h_\theta$ is the batch-level representation histogram. %$h_\theta$ denotes the batch-level histogram from the current generalist model.

%This formulation reduces MD-MTL to a problem of \emph{distributional alignment in representation space}: given a target distribution $\mu_u^{(\ell)}$ and a collection of source distributions $\{\mu_{m,t}^{(\ell)}\}$, we seek a principled mechanism to measure their similarity, select or weight relevant sources, and aggregate them into a unified representation suitable for learning a generalist model.

\section{Experiments}
\subsection{Experiment Settings}
\subsubsection{Datasets}

We evaluate our proposed model on four representative NLP tasks: Paraphrasing, Summarization, Question Answering, and Information Extraction. Each task covers four distinct domains, as summarized in Table~\ref{tab:dataset}, which lists the corresponding datasets used for each domain. 
In total, our experiments involve 30 publicly available datasets for training and evaluation.
To ensure fair comparison and reproducibility, we follow the same training, validation, and testing splits provided by each dataset. 
Additionally, we construct a probing set for each task and domain by randomly sampling 2k instances from the original training set, ensuring that the probing data do not overlap with any examples used for training, validation, or testing.
We further sample 5k data from the training set for each task and domain for the generalist training. 
%Detailed description of all datasets are included in Sec.~\ref{sec:data_appendix} in the Appendix.

% Paraphrase: biology: elife + plos + cells; news (SciTechNews + Microsoft Research Paraphrase Corpus paper), science: vtech + parasci (ACL + Arxiv), wiki: paws
% QA datasets: web: ComplexWebQuestions + WebQuestions + triviaQA, biomedical: COVID-QA + healthcaremagic, Wiki: wikiQA + natural questions, news: newsQA
% Summarization: news: 
%https://lil.nlp.cornell.edu/newsroom/index.html
%https://huggingface.co/datasets/alexfabbri/multi_news
%https://www.kaggle.com/datasets/datagator/wikinews-article-dataset
%https://github.com/complementizer/wcep-mds-dataset
% literatures: booksum + SQuALITY, goverment: billsum + bigpatent + GoVReport; science: https://github.com/armancohan/long-summarization/tree/master?tab=readme-ov-file (arxiv+pubmed) + https://www.kaggle.com/datasets/jawakar/scisummnet-corpus + bigsuvery + multi_xscience + SumPumed
% IE: IEPILE. News: ["DuIE2.0", "NYT", "NYT11-HRL", "CoNLL2004", "KBP37", "GIDS", "CLUE", "MSRA", "ACE2005", "CoNLL2003", "DuEE1.0"], bio: ["CMeIE", "ADE Corpus", "PHEE", "AnatEM", "BC2GM", "BC4CHEMD", "BC5CDR", "NCBI-Disease", "GENIA"], sci: ["Semeval RE", "SciERC", "FabNER", "CASIE"], wiki: ["MultiNERD"]

\subsubsection{Baselines}
We compare HWM against several representative baselines.
These include the pretrained base model (LLaMA3.1-8B-Instruct~\cite{dubey2024llama}), specialists (models fine-tuned based only on individual domain-task datasets), and a generalist model fine-tuned jointly on the union of all domain-task data.
For model merging approaches, we consider a range of general model merging methods, including Model Soup~\cite{wortsman2022model}, Task Arithmetic~\cite{ilharco2023editing}, TIES~\cite{yadav2023ties}, DARE~\cite{yu2024language}, DELLA~\cite{deep2024della}, Model Breadcrumbs~\cite{davari2024model}, and Model Stock~\cite{jang2024model}. %All merging baselines are implemented using MergeKit~\footnote{\url{https://github.com/arcee-ai/mergekit}}.
For our approach, we evaluate two variants of Hierarchical Wasserstein Merging: (i) HWM, a non-training-based version that aggregates specialists using Wasserstein-derived hierarchical weights, and (ii) $\text{HWM}^{*}$, a training-based single generalist learned through a hybrid Wasserstein loss.
%supervised fine-tuning augmented with a Wasserstein alignment loss.

\begin{table*}[t]
\centering
\caption{Performance (\%) comparison of merging methods for text summarization across four domains (Government, Literature, News, and Science). We report BERTScore F1 (\%), SARI, and human-aligned preference scores using LLM-as-a-judge. The overall best result is in bold, and the best merging result is underlined.}
\scriptsize
\begin{tabular}{ll|cccc|cccc|cccc}
\toprule
Types & Methods
 & \multicolumn{4}{c}{BERTScore F1 (\%)} 
 & \multicolumn{4}{c}{SARI} 
 & \multicolumn{4}{c}{LLM-as-a-judge (\%)} \\
\midrule
 & 
 & Gov. & Lit. & News & Sci.
 & Gov. & Lit. & News & Sci.
 & Gov. & Lit. & News & Sci. \\
\midrule

Pre-trained
 & Pretrained
 & 74.92 & 73.18 & 71.40 & 72.63
 & 35.21 & 36.08 & 36.44 & 34.97
 & 52.34 & 42.10 & 62.93 & 72.53 \\

\midrule

Fine-tuned
 & Generalist
 & 81.68 & 76.22 & 74.33 & 80.49
 & 44.88 & 39.22 & 37.40 & 44.76
 & 51.76 & 39.87 & 57.56 & 63.54 \\

 & Specialist
 & \textbf{82.12} & 77.25 & 76.58 & \textbf{81.27}
 & \textbf{45.56} & 39.70 & 41.50 & \textbf{45.27}
 & 53.26 & 44.14 & 65.69 & 69.84 \\

\midrule

Merged
 & Model Soup
 & 76.81 & 76.49 & 75.39 & 76.93
 & 40.32 & 37.47 & 35.33 & 38.52
 & 49.96 & 44.38 & 59.20 & 63.88 \\

 & Task Arithmetic
 & 78.43 & 76.72 & 76.30 & 75.87
 & 38.53 & 39.16 & 38.78 & 37.78
 & 64.22 & 50.46 & 60.51 & 72.27 \\

 & TIES
 & 78.03 & 74.41 & 74.34 & 75.20
 & 37.48 & 38.07 & 40.44 & 37.74
 & 58.98 & 35.49 & 48.57 & 65.26 \\

 & DARE
 & 77.66 & 74.75 & 75.75 & 75.40
 & 38.13 & 38.18 & 38.59 & 37.79
 & 59.14 & 35.44 & 54.46 & 66.90 \\

 & DELLA
 & 78.02 & 74.41 & 74.74 & 75.21
 & 38.26 & 38.07 & 40.29 & 38.63
 & 58.10 & 35.66 & 50.70 & 67.02 \\

 & Breadcrumbs
 & 77.03 & 76.31 & 75.25 & 75.67
 & 37.23 & 38.94 & 40.23 & 37.04
 & 64.37 & 48.32 & 59.39 & 72.09 \\

 & Model Stock
 & 65.93 & 66.02 & 64.68 & 64.53
 & 33.78 & 37.88 & 38.36 & 35.65
 & 59.24 & 35.07 & 46.60 & 56.38 \\

 & \textbf{HWM (ours)}
 & \underline{79.94}  & 76.53  & 77.30 & 77.75
 & \underline{41.15} & 39.54 & 42.09 & 39.61
 & \textbf{\underline{67.28}} & \textbf{\underline{53.32}} & 60.79  & 78.40  \\

 & \textbf{HWM$^{*}$ (ours)}
 & 77.28 & \textbf{\underline{77.41}} & \textbf{\underline{77.37}} & \underline{79.48}
 & 39.48 & \textbf{\underline{39.96}}& \textbf{\underline{42.24}} & \underline{40.21}
 & 63.00 & 53.23 & \textbf{\underline{66.13}} & \textbf{\underline{80.90}} \\

\bottomrule
\end{tabular}

\label{tab:summarization}
\end{table*}

\subsubsection{Evaluation Metrics}
We adopt task-specific evaluation metrics to assess model performance across different tasks. Overall, our evaluation metrics consist of two categories: automative model assessment and LLM-as-a-judge, where GPT4.1 is used to
score the candidate outputs given the corresponding input context and task instructions~\cite{li2025generation}. LLM-as-a-judge evaluation is applied to all four tasks~\footnote{Due to the cost of GPT4.1, we randomly sample 1k testing data per domain of each task for LLM-as-a-judge assessment.}, and the detailed prompts are provided in Appendix~\ref{sec:prompt}.
For text summarization and lay paraphrasing, we report BERTScore (F1)~\cite{Zhang*2020BERTScore:}, which measures semantic similarity between generated and reference texts, and SARI~\cite{xu-etal-2016-optimizing}, which evaluates content selection and rewriting behavior.
For the generative question answering task, we use BERTScore (F1)~\cite{Zhang*2020BERTScore:} and METEOR~\cite{banerjee-lavie-2005-meteor}, as METEOR better captures semantic matches in generative QA and is more suitable for abstractive responses.
For information extraction, we report the prediction validity rate, defined as the percentage of model outputs that can be parsed into a valid dictionary, to measure format reliability. We also report Micro-F1, computed as the micro-averaged extraction F1 score over canonicalized outputs using soft matching, to assess extraction accuracy.

\begin{table*}[t]
\centering
\caption{Performance (\%) comparison of merging methods for lay paraphrasing across four domains (Biology, News, Science, and Wikipedia). We report BERTScore F1 (\%), SARI, and human-aligned preference scores using LLM-as-a-judge. The overall best result is in bold, and the best merging result is underlined.}
\scriptsize
\begin{tabular}{ll|cccc|cccc|cccc}
\toprule
Types & Methods
 & \multicolumn{4}{c}{BERTScore F1 (\%)} 
 & \multicolumn{4}{c}{SARI} 
 & \multicolumn{4}{c}{LLM-as-a-judge (\%)} \\
\midrule
 & 
 & Bio. & News & Sci. & Wiki
 & Bio. & News & Sci. & Wiki
 & Bio. & News & Sci. & Wiki \\
\midrule

Pre-trained
 & Pretrained
 & 70.16 & 61.20 & 62.81 & 69.25
 & 38.24 & 23.80 & 25.65 & 28.23
 & 54.38 & 10.26 & 9.19 & 24.71 \\

\midrule

Fine-tuned
 & Generalist
 & 77.18 & 69.85 & 74.58 & 78.15
 & 38.75 & 40.26 & 34.48 & 42.44
 & 55.28 & 14.59 & 10.09 & 20.03 \\

 & Specialist
 & 81.41 & 74.08 & \textbf{81.19} & \textbf{90.60}
 & 41.46 & \textbf{44.99} & \textbf{39.33} & \textbf{51.79}
 & 61.61 & 21.21 & 33.51 & \textbf{65.06} \\

\midrule

Merged
 & Model Soup
 & 81.08 & 76.64 & 78.88 & 84.34
 & 39.36 & 36.43 & 34.80 & 34.84
 & 53.77 & 25.36 & 15.17 & 40.86 \\

 & Task Arithmetic
 & 80.40 & 74.59 & 75.45 & 75.20
 & 38.70 & 33.86 & \underline{34.89} & 33.86
 & 64.56 & 23.52 & 26.61 & 21.07 \\

 & TIES
 & 80.46 & 78.21 & 78.88 & \underline{84.75}
 & 38.45 & 33.81 & 33.42 & 21.99
 & 69.91 & 39.52 & 22.48 & 40.28 \\

 & DARE
 & 81.10 & 78.21 & 77.56 & 78.19
 & 38.72 & 33.81 & 34.52 & 34.81
 & 68.46 & 42.86 & 39.71 & 33.02 \\

 & DELLA
 & 80.47 & 78.59 & 79.09 & 76.43
 & 38.50 & 33.79 & 33.41 & 35.03
 & 66.19 & 41.33 & 31.41 & 22.21 \\

 & Breadcrumbs
 & 80.58 & 74.59 & 80.26 & 80.93
 & 39.92 & 33.86 & 30.86 & 21.57
 & 64.40 & 23.42 & 25.62 & 21.10 \\

 & Model Stock
 & 65.98 & 59.52 & 65.77 & 62.56
 & 34.26 & 26.90 & 28.52 & 27.52
 & 52.70 & 22.81 & 13.44 & 40.70 \\

 & \textbf{HWM (ours)}
 & 81.74  & 79.27 & \underline{80.65}  & 83.33
 & 40.19 & 38.92 & 34.52 & \underline{36.87}
 & \textbf{\underline{70.14}}  & 43.49 & \textbf{\underline{43.71}} & 44.77 \\

 & \textbf{HWM$^{*}$ (ours)}
 & \textbf{\underline{83.48}} & \textbf{\underline{79.84}} & 79.12 & 77.70
 & \textbf{\underline{42.48}} & \underline{40.97} & 32.39 & 33.93
 & 64.37 & \textbf{\underline{45.06}} & 36.49 & \underline{46.48} \\

\bottomrule
\end{tabular}
\label{tab:lay}
\end{table*}

\begin{table*}[t]
\centering
\caption{Performance (\%) comparison of merging methods for generative question answering across four domains (Biology, News, Web, and Wikipedia). We report BERTScore F1 (\%), METEOR (\%), and human-aligned preference scores using LLM-as-a-judge. The overall best result is in bold, and the best merging result is underlined.}
\scriptsize
\begin{tabular}{ll|cccc|cccc|cccc}
\toprule
Types & Methods
 & \multicolumn{4}{c}{BERTScore F1 (\%)} 
 & \multicolumn{4}{c}{METEOR (\%)} 
 & \multicolumn{4}{c}{LLM-as-a-judge (\%)} \\
\midrule
 & 
 & Bio. & News & Web & Wiki
 & Bio. & News & Web & Wiki
 & Bio. & News & Web & Wiki \\
\midrule

Pre-trained
 & Pretrained
 & 67.66 & 72.92 & 67.51 & 67.36
 & 12.49 & 12.54 & 20.32 & 15.85
 & 25.67 & 25.02 & 58.70 & 70.87 \\

\midrule

Fine-tuned
 & Generalist
 & 72.54 & 77.53 & 85.40 & 69.74
 & 18.69 & 31.26 & 18.31 & 15.38
 & 48.15 & 41.15 & 59.57 & 61.50 \\

 & Specialist
 & 77.43 & 77.77 & \textbf{85.66} & 72.04
 & 22.17 & \textbf{32.01} & \textbf{36.25} & \textbf{28.16}
 & 64.83 & 47.17 & 68.11 & 74.85 \\

\midrule

Merged
 & Model Soup
 & 74.34 & 77.50 & 72.46 & 75.85
 & 11.57 & 30.81 & 23.40 & 23.03
 & 46.26 & 43.47 & 58.68 & 66.37 \\

 & Task Arithmetic
 & 75.12 & 75.56 & 73.50 & 76.80
 & 16.49 & 23.49 & 27.71 & \underline{27.65}
 & 63.02 & \textbf{\underline{50.45}} & \textbf{\underline{68.22}} & 79.57 \\

 & TIES
 & 76.49 & 76.09 & 74.32 & 77.46
 & 17.83 & 25.52 & 27.72 & 26.27
 & 51.10 & 41.79 & 55.13 & 69.74 \\

 & DARE
 & 74.39 & 76.16 & 73.36 & 77.45
 & 12.28 & 25.77 & 27.25 & 27.15
 & 54.46 & 44.77 & 63.29 & 79.40 \\

& DELLA
 & 75.78 & 77.09 & 75.27 & 77.49
 & 18.46 & 25.60 & 27.72 & 26.31
 & 59.93 & 42.86 & 55.36 & 69.30 \\

 & Breadcrumbs
 & 75.88 & 76.67 & 73.50 & 76.09
 & \textbf{\underline{22.24}} & 26.86 & 27.71 & 26.28
 & \textbf{\underline{68.37}} & 49.19 & 67.90 & 79.86 \\

 & Model Stock
 & 74.43 & 77.42 & 71.24 & 77.65
 & 11.47 & 31.07 & 22.89 & 27.59
 & 56.48 & 45.12 & 61.77 & \textbf{81.64} \\

 & \textbf{HWM (ours)}
 & \textbf{\underline{77.82}} & 77.42 & 77.80  & \textbf{\underline{78.30}}
 & 21.23 & 28.21 & 28.45 & 27.11
 & 67.17 & 45.40 & 62.62 & 79.69 \\

 & \textbf{HWM$^{*}$ (ours)}
 & 77.64 & \textbf{\underline{80.39}} & \underline{79.54} & 78.00
 & 19.93 & \underline{31.50} & \underline{30.91} & \underline{27.65}
 & 65.13 & 49.51 & 67.19 & 81.18 \\

\bottomrule
\end{tabular}
\label{tab:QA}
\end{table*}

\begin{table*}[t]
\centering
\caption{Performance (\%) comparison for information extraction across four domains (Biology, News, Science, and Wikipedia). We report output validity rate (\%), Micro-F1 (\%) for extraction accuracy, and human-aligned preference scores using LLM-as-a-judge. The overall best result is in bold, and the best merging result is underlined.}
\scriptsize
\begin{tabular}{ll|cccc|cccc|cccc}
\toprule
Types & Methods
 & \multicolumn{4}{c}{Prediction Validity Rate (\%)} 
 & \multicolumn{4}{c}{Micro-F1 (\%)} 
 & \multicolumn{4}{c}{LLM-as-a-judge (\%)} \\
\midrule
 & 
 & Bio. & News & Sci. & Wiki
 & Bio. & News & Sci. & Wiki
 & Bio. & News & Sci. & Wiki \\
\midrule

Pre-trained
 & Pretrained
 & 16.22 & 25.70 & 26.37 & 28.98
 & 5.84 & 20.80 & 5.36 & 22.85
 & 11.60 & 24.38 & 14.07 & 60.90 \\

\midrule

Fine-tuned
 & Generalist
 & 3.55 & 8.20 & 8.16 & 15.36
 & 0.44 & 8.69 & 7.10 & 19.02
 & 24.70 & 17.66 & 15.32 & 59.15 \\

 & Specialist
 & 47.19 & 92.67 & 55.02 & 84.00
 & \textbf{27.17} & 31.01 & 11.40 & 33.33
 & \textbf{31.42} & 44.05 & 17.66 & 62.40 \\

\midrule

Merged
 & Model Soup
 & 25.08 & 16.05 & 22.29 & 16.45
 & 9.36 & 16.07 & 5.70 & 18.17
 & 15.36 & 15.18 & 15.40 & 53.16 \\

 & Task Arithmetic
 & 14.85 & 97.65 & 17.99 & \textbf{\underline{97.71}}
 & 5.37 & 32.84 & 6.87 & 37.65
 & 19.81 & 35.92 & 13.64 & 67.50 \\

 & TIES
 & 81.25 & 93.49 & 75.57 & 86.71
 & 13.32 & 29.32 & 12.37 & 36.10
 & 13.85 & 29.02 & 12.97 & 62.75 \\

 & DARE
 & 30.03 & 90.30 & 77.83 & 87.25
 & 8.67 & 27.32 & 12.52 & 31.04
 & 20.50 & 28.37 & 14.42 & 64.15 \\

 & DELLA
 & 81.58 & 90.35 & 26.41 & 87.36
 & 11.49 & 27.35 & 9.21 & 31.11
 & 17.67 & 28.05 & 12.24 & 64.54 \\

 & Breadcrumbs
 & \textbf{\underline{91.53}} & 98.05 & \textbf{\underline{94.32}} & 97.71
 & \underline{13.43} & 32.60 & \textbf{\underline{15.42}} & 37.68
 & 14.58 & 36.81 & 13.42 & \textbf{\underline{71.47}} \\

 & Model Stock
 & 20.18 & 21.25 & 17.74 & 22.77
 & 7.58 & 14.47 & 3.61 & 17.56
 & 14.92 & 22.59 & 12.76 & 60.27 \\

 & \textbf{HWM (ours)}
 & 85.71  & 98.25  & 94.10 & 94.12 
 & 12.34 & 36.60 & 15.27 & 37.14
 & 22.90 & 44.87 & 20.48 & 69.27 \\

 & \textbf{HWM$^{*}$ (ours)}
 & 85.66 & \textbf{\underline{98.90}} & 92.20 & 95.56
 & 11.81 & \textbf{\underline{41.32}} & 15.38 & \textbf{\underline{39.66}}
 & \underline{23.65} & \textbf{\underline{47.22}} & \textbf{\underline{22.98}} & 70.39 \\

\bottomrule
\end{tabular}
\label{tab:IE}
\end{table*}

\subsubsection{Parameter Settings}
We implement Hierarchical Wasserstein Merging (HWM) in Pytorch and finetune both specialists and generalist models using LLaMA-3.1-8B-Instruct~\cite{dubey2024llama} by LLaMA-Factory~\footnote{\url{https://github.com/hiyouga/LlamaFactory}} as the shared pretrained backbone. For specialists training, we use the learning rate of 1e-5 and train each specialist for one epoch on its corresponding domain-task dataset $\mathcal{D}_{t,m}$. 
For Wasserstein aggregation, we sample up to 2k probing examples per dataset. %Details of parameter settings for HWM are in Appendix~\ref{sec:appendix_setting}.

The number of cluster centroids $K$ is selected by grid search from $\{128,256,512\}$. Histogram smoothness $\tau$ is 1.2 in equation~\ref{equ:kmeans}. 
The entropic regularization $\varepsilon$ is 0.07 for building the kernel $K_{ij}$. Barycenters run for up to 200 iterations with a tolerance of $10^{-9}$. 
For the non-training hierarchical weighting in Equ.~\ref{equ:weights}, $\beta_{task}$ and $\beta_{in}$ are 0.1 and 0.5, respectively. 
For generalist training, both $\lambda_{task}$ and $\lambda_{global}$ are 0.3 in Equ.~\ref{equ:training}. The Sinkhorn solver for the alignment loss runs for 10 iterations, and the WAL loss is applied periodically every two optimization steps.
All merging baselines are performed using mergekit~\footnote{\url{https://github.com/arcee-ai/mergekit}}.

\subsection{Results and Analysis}
\subsubsection{Main Results}

\paragraph{Text Summarization.} Table~\ref{tab:summarization} reports summarization results across four domains
using BERTScore F1, SARI, and LLM-as-a-judge. HWM achieves the strongest overall performance among merging methods, obtaining the best or second-best results on BERTScore and SARI in the Literature, News, and Science domains. It also substantially improves LLM-as-a-judge scores, especially in Science, suggesting that representation-level Wasserstein aggregation better preserves summary quality aligned with human preferences. While fully fine-tuned generalists or specialists can be competitive in data-rich domains such as Government and Science, HWM remains robust across domains and shows clear advantages under greater domain heterogeneity.

\paragraph{Lay Paraphrasing.} Table~\ref{tab:lay} reports lay paraphrasing results across four domains using BERTScore F1, SARI, and LLM-as-a-judge. Among merging-based approaches, HWM consistently achieves the strongest overall performance, getting the best or second-best results across most domains and metrics. In particular, HWM yields the highest BERTScore and SARI scores in the Biomedical and News domains, indicating improved semantic preservation and paraphrasing quality under domain shift. 
HWM also outperforms other merging baselines on LLM-as-a-judge scores in News and Wikipedia, suggesting better alignment with human preferences. While fully fine-tuned specialists are very competitive, especially in high-resource domains (i.e., Wikipedia), HWM substantially narrows the performance gap without requiring domain-specific inference, demonstrating its effectiveness as a generalist merging framework for lay paraphrasing.

\begin{table*}[t]
\centering
\caption{Ablation study of HWM components. Results are averaged across domains for each task and reported in percentages. BS denotes BERTScore F1, MET. denotes METEOR, Valid. denotes prediction validity rate, MiF1 denotes Micro-F1, and LLM denotes LLM-as-a-judge.}
\scriptsize
\begin{tabular}{ll|ccc|ccc|ccc|ccc}
\toprule
Types & Methods
 & \multicolumn{3}{c}{Summarization}
 & \multicolumn{3}{c}{Lay}
 & \multicolumn{3}{c}{QA}
 & \multicolumn{3}{c}{IE} \\
\midrule
 & 
 & BS & SARI & LLM
 & BS & SARI & LLM
 & BS & MET. & LLM
 & Valid. & MiF1 & LLM \\
\midrule

Architecture
 & Global
 & 74.98 & 37.03 & 57.31
 & 80.89 & 34.46 & 46.76
 & 76.31 & 24.11 & 60.62
 & 76.38 & 15.88 & 22.35 \\

 & HWM
 & \textbf{77.88} & \textbf{40.60} & \textbf{64.95}
 & \textbf{81.25} & \textbf{37.63} & \textbf{50.53}
 & \textbf{77.84} & \textbf{26.25} & \textbf{63.72}
 & \textbf{93.05} & \textbf{25.34} & \textbf{39.38} \\

\midrule

Centroids
 & HWM-128
 & 75.28 & 38.50 & 51.27
 & 80.35 & 34.00 & 47.88
 & 76.01 & 38.64 & 60.47
 & 89.25 & 19.95 & 34.12 \\

 & HWM-256
 & 76.58 & 39.68 & 57.08
 & 80.89 & 33.92 & 48.46
 & 76.38 & \textbf{38.66} & 59.88
 & 92.09 & 20.01 & 35.87 \\

 & HWM-512
 & 77.88 & \textbf{40.60} & 64.95
 & \textbf{81.25} & \textbf{37.63} & \textbf{50.53}
 & 77.84 & 26.25 & 63.72
 & 93.05 & 25.34 & 39.38 \\

 & HWM$^{*}$-128
 & 73.14 & 36.57 & 60.68
 & 78.71 & 35.35 & 43.35
 & 76.73 & 37.39 & 60.12
 & 90.48 & 23.54 & 39.63 \\

 & HWM$^{*}$-256
 & 75.56 & 38.29 & 61.87
 & 79.04 & 36.48 & 45.50
 & 77.73 & 37.86 & 64.12
 & 92.50 & 25.81 & 37.93 \\

 & HWM$^{*}$-512
 & \textbf{77.89} & 40.47 & \textbf{65.82}
 & 80.04 & 37.44 & 48.10
 & \textbf{78.89} & 27.50 & \textbf{65.75}
 & \textbf{93.08} & \textbf{27.04} & \textbf{41.06} \\

\midrule

Probing
 & HWM-small
 & 75.13 & 38.40 & 60.48
 & 79.60 & 34.30 & 48.68
 & 75.39 & 24.12 & 56.61
 & 87.27 & 20.97 & 34.94 \\

 & HWM
 & 77.88 & \textbf{40.60} & 64.95
 & \textbf{81.25} & \textbf{37.63} & \textbf{50.53}
 & 77.84 & 26.25 & 63.72
 & 93.05 & 25.34 & 39.38 \\

 & HWM$^{*}$-small
 & 75.88 & 38.46 & 64.77
 & 79.75 & 35.20 & 45.92
 & 78.47 & 23.89 & 63.79
 & 90.05 & 25.86 & 39.60 \\

 & HWM$^{*}$
 & \textbf{77.89} & 40.47 & \textbf{65.82}
 & 80.04 & 37.44 & 48.10
 & \textbf{78.89} & \textbf{27.50} & \textbf{65.75}
 & \textbf{93.08} & \textbf{27.04} & \textbf{41.06} \\

\bottomrule
\end{tabular}
\label{tab:ablation}
\end{table*}

\paragraph{Generative Question Answering.}
Table~\ref{tab:QA} summarizes generative question answering performance across four domains using BERTScore F1, METEOR, and LLM-as-a-judge. Among merging-based approaches, HWM achieves the strongest and most consistent performance across domains, obtaining the best merging results on BERTScore in all four domains and competitive METEOR scores, particularly in Biology, News, and Wikipedia. %HWM also performs favorably on LLM-as-a-judge, indicating strong alignment with human preferences while maintaining robust semantic fidelity. 
In addition, we observe that in Web and Wikipedia domains, the pretrained model already exhibits strong performance, likely due to broad exposure to similar content during pretraining. As a result, additional fine-tuning or merging yields more modest gains in these high-resource domains, while HWM shows clearer improvements in domains with greater distributional divergence.
%Although fully fine-tuned specialists remain competitive—especially in high-resource settings such as Web QA—HWM substantially narrows the gap without relying on domain-specific inference, demonstrating its effectiveness as a generalist merging framework for generative question answering.

\paragraph{Information Extraction.}
Table~\ref{tab:IE} presents information extraction results across four domains using prediction validity rate, Micro-F1, and LLM-as-a-judge. 
From Table~\ref{tab:IE}, we observe that the jointly trained generalist performs poorly on information extraction, likely due to interference from heterogeneous tasks that disrupt the strict output structure required by IE. This behavior is similar to the pretrained model, which often fails to produce valid structured outputs, highlighting the challenge of preserving task-specific output formats under joint training.
%HWM achieves strong performance particularly in high-resource domains such as News and Wikipedia. 
%HWM attains the best merging results on Micro-F1 in News and Wikipedia, indicating improved extraction accuracy while maintaining high validity rates. 
Compared to other merging baselines that often exhibit unstable behavior across domains, HWM provides a more balanced performance profile, yielding competitive validity, higher Micro-F1, and strong human-aligned preference scores. 

%While task-specific specialists remain strongest in certain domains, HWM substantially narrows this gap without requiring domain-specific inference, demonstrating its effectiveness as a generalist merging framework for structured prediction tasks.

\subsubsection{Ablation Study}
To evaluate key design choices in HWM, we conduct ablation studies on the hierarchical architecture, the number of cluster centroids, and the probing-set size. Table~\ref{tab:ablation} reports domain-averaged scores for each task. We observe that each component contributes positively. The hierarchical architecture consistently outperforms the global variant, with particularly large gains on the IE task.
Increasing the number of centroids generally improves performance, with $K=512$ achieving the best or near-best results in most cases. The full probing setup also outperforms the smaller probing set, especially on QA and IE tasks. These results support our choices of hierarchical aggregation, $K=512$, and using the full probing set.
%We report the ablation results of each domain in the Appendix~\ref{sec:appendix_ablation}.

\paragraph{Effect of hierarchical architecture.}
Table~\ref{tab:appendix_ablation_arch} in the Appendix shows that the hierarchical architecture improves over the global architecture in nearly all domains and metrics. For summarization, HWM improves BERTScore and LLM-as-a-judge scores across all four domains, with especially large gains on the Science domain. Similar trends are observed for Lay and QA, where HWM improves most domain-level scores despite a few small decreases on individual metrics. The improvement is most consistent on IE where HWM outperforms the global architecture in every domain for prediction validity, Micro-F1, and LLM-as-a-judge scores. These results suggest that hierarchical aggregation better preserves domain-specific information before global fusion, leading to more robust performance across tasks and domains.

\paragraph{Effect of the number of centroids.}
Table~\ref{tab:appendix_ablation_centroids} in the Appendix reports the effect of varying the number of cluster centroids $K$. Overall, increasing $K$ generally improves performance for both HWM and HWM$^\ast$, with $K=512$ achieving the best or near-best results in most domains. The gains are particularly clear for summarization and IE, where larger centroid sets improve most metrics across domains. Although a few individual domain-metric pairs show non-monotonic trends, the overall pattern indicates that finer centroid granularity helps capture more task- and domain-specific structure. This supports our choice of using $K=512$ as the default setting.

\paragraph{Effect of probing size.}
Table~\ref{tab:appendix_ablation_probing} in Appendix compares the full probing setup with a smaller probing set. The full probing setup consistently improves HWM across most domains, especially for Summarization, Lay, and IE. For IE, using the full probing set improves prediction validity and Micro-F1 in every domain for both HWM and HWM$^\ast$. QA also benefits from more probing examples, particularly in LLM-as-a-judge scores. While a small number of domain-level metrics decrease, the overall trend shows that additional probing examples provide more reliable task- and domain-level signals, leading to stronger model merging performance.

\subsubsection{Additional Analysis}
\paragraph{Non-training-based v.s. Training-based HWM}
Across all four tasks, both training (HWM$^{*}$) and non-training (HWM) versions of our proposed model demonstrate strong performance as shown from Table~\ref{tab:summarization} to Table~\ref{tab:IE}. HWM consistently improves over existing training-free merging baselines, achieving competitive or best merging performance on semantic metrics such as BERTScore and LLM-as-a-judge scores, while requiring no additional optimization. 
HWM$^{*}$ further enhances performance by aligning the generalist’s representations with task-level and global Wasserstein barycenters, leading to additional gains on metrics such as SARI for summarization and Micro-F1 for information extraction. In particular, HWM$^{*}$ shows clear advantages in domains where preserving task-specific structure is critical, while HWM remains robust and effective in scenarios where retraining is infeasible. These results highlight the flexibility of HWM, enabling both efficient training-free aggregation and performance-oriented generalist learning within a unified framework.

\begin{figure}[t]
    \centering
    \includegraphics[width=\linewidth]{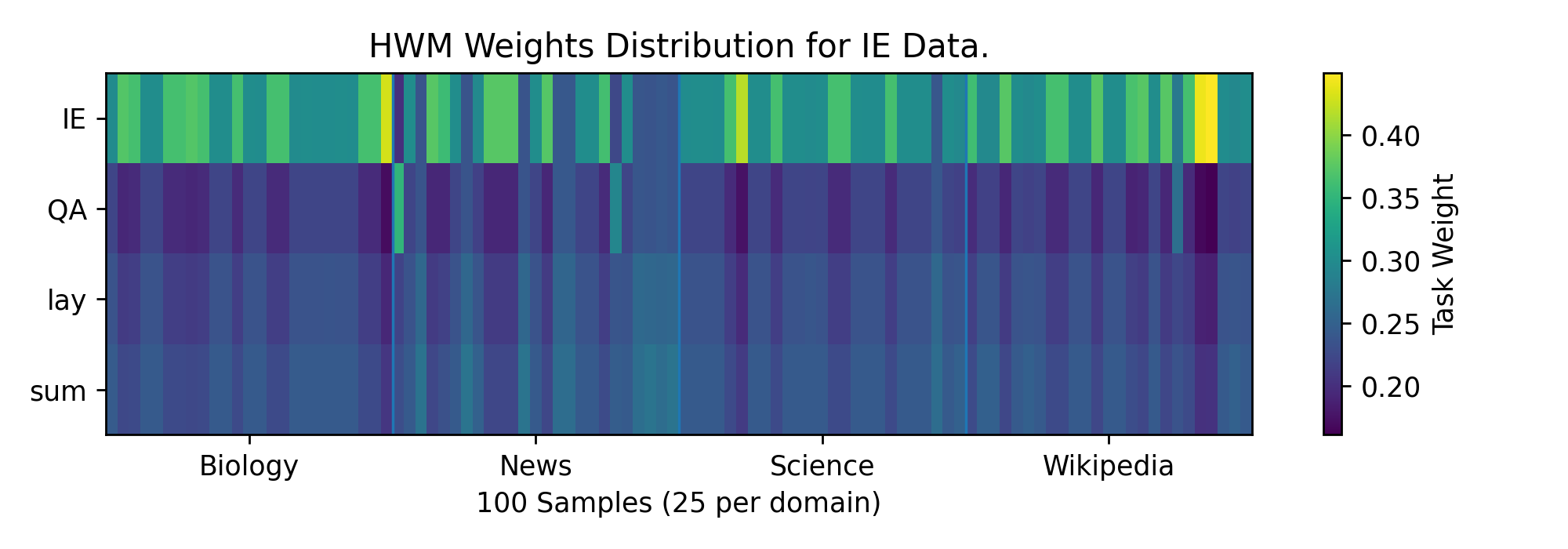}

    \vspace{0.5em}

    \includegraphics[width=\linewidth]{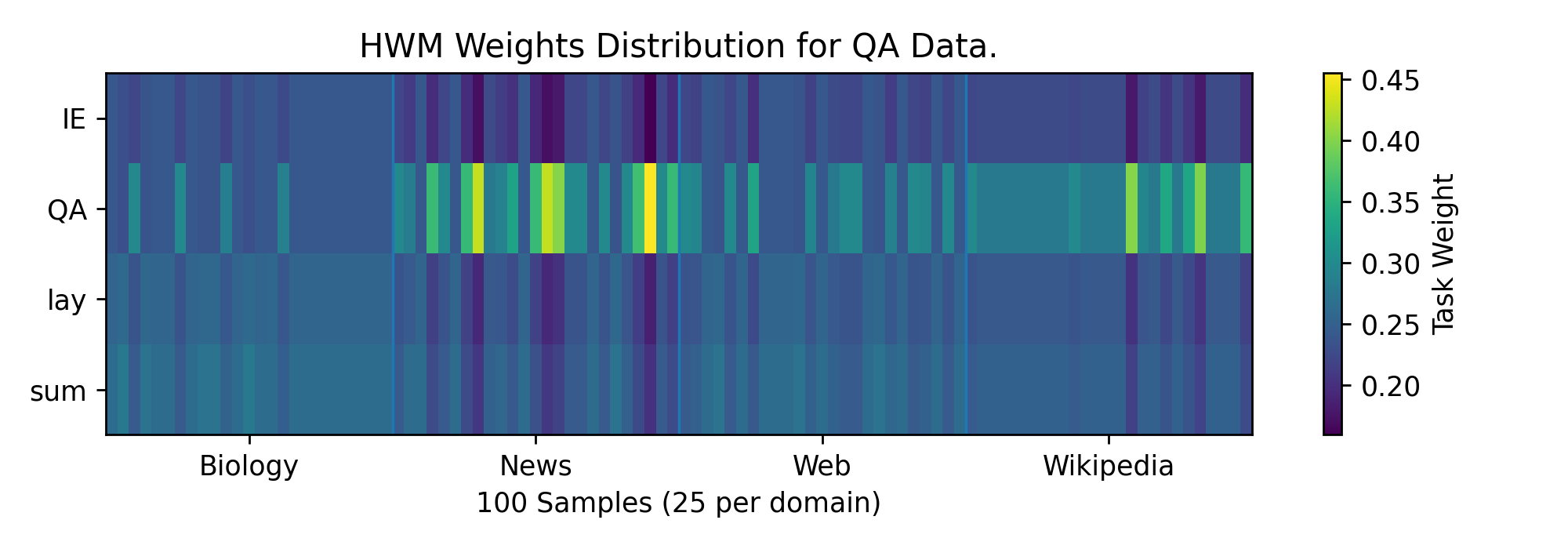}

    \caption{Visualization of HWM weights for IE and QA tasks. Each column is one data sample, and each row shows the average weight assigned to specialists from different tasks. Color intensity indicates the magnitude of the assigned task weight.}
    \label{fig:distribution}
\end{figure}

\paragraph{Task Weight Distributions.}
Figure~\ref{fig:distribution} visualizes the task-level weight distributions produced by HWM for IE and QA inputs across multiple domains. For IE inputs, HWM consistently assigns higher weights to IE specialists, while still allocating weights to related tasks such as QA, reflecting shared factual and structural information. 
Similarly, for QA inputs, QA specialists receive dominant weights. 
These results demonstrate that HWM does not rely on hard task assignments.
Instead, it performs fine-grained, sample-dependent weighting that captures both task alignment and cross-task relatedness, supporting its effectiveness as a flexible multi-domain multi-task merging framework.

\section{Conclusion}
In this paper, we propose Hierarchical Wasserstein Merging (HWM), a unified framework for multi-domain multi-task learning. 
Without requiring domain and task-specific information during inference, HWM supports both non-training-based specialist aggregation through Wasserstein-derived weights and training-based generalist learning through a hybrid supervised and alignment loss.
%Specifically, 
We evaluate HWM against multiple model merging baselines across four tasks and four domains per task under diverse evaluation metrics.
Experimental results demonstrate the effectiveness of HWM by showing that the generalist can be performant simultaneously in multi-domain multi-task learning.
In future work, we plan to extend HWM to multimodal architectures and broader task settings.

\section{Limitations}
\paragraph{Limitations.}
Our work has several limitations that suggest directions for future research.
First, we compare HWM with a range of representative model merging methods with publicly available implementations. However, several recent approaches do not release code, preventing direct empirical comparison. We also exclude methods designed only for merging two models or methods requiring additional supervision signals, such as user preferences, to ensure a fair comparison under our multi-model unsupervised merging setting.

Second, due to the cost of LLM-based evaluation, we apply the GPT-4.1 LLM-as-a-judge protocol to a randomly sampled subset of 1k test instances per domain and task. This results in 16k judged examples, covering approximately 8\% of the full test set of over 191k instances. While this provides a substantial human-aligned evaluation, sampling may introduce variance and reduce statistical confidence compared with full-test evaluation. %Evaluating all main models on the full test set would cost approximately \$36k, exceeding our budget. 
To mitigate this limitation, we evaluate all models on the full test set using automatic metrics, including BERTScore, SARI, METEOR, prediction validity, and Micro-F1, and observe trends consistent with the GPT-4.1 judgments.

Third, our experiments are conducted with a single backbone, LLaMA-3.1-8B-Instruct. Fixing the backbone allows us to isolate the effect of different merging strategies under controlled conditions, but it leaves open how well HWM generalizes to other model families and scales. Extending HWM to more backbones and incorporating additional recent merging methods are important directions for future work.
%Evaluating HWM with other backbones remains an important di for future investigation.


\bibliography{custom}

\appendix

\section{Appendix}

\subsection{Coordinate consistency of the shared support.}
HWM assumes that the specialists share a common representation coordinate system, which holds in our setting because all specialists are fine-tuned from the same pretrained backbone and have the same architecture. The shared support $C=\{c_1,\ldots,c_K\}$ is not constructed separately for each specialist; instead, we pool representations from all specialists and fit K-means once on this union. This yields a single reference partition of the shared representation space. Moreover, the subsequent soft histograms and Wasserstein costs depend only on Euclidean distances to the shared centroids. Therefore, a global relabeling or permutation of representation coordinates applied consistently to all specialists would leave the distances, cluster assignments, and optimal transport costs unchanged. In contrast, arbitrary model-specific coordinate permutations would require an additional alignment step. Such permutations are unlikely in our experimental setup because all specialists are initialized from the same backbone rather than trained independently from scratch.

\subsection{Dataset Comparisons} \label{sec:dataset_description}
Table~\ref{tab:dataset} summarizes the datasets used in our experiments across four NLP tasks and multiple domains. We construct a diverse evaluation suite covering paraphrasing, question answering, summarization, and information extraction. For each task, we include datasets from several domains, such as biomedical, news, science, Wikipedia, government, literature, and web sources. This design allows us to evaluate whether model merging methods can preserve both task-specific and domain-specific capabilities.

For paraphrasing, we include datasets from biomedical, news, science, and Wikipedia domains. For question answering, we cover web, biomedical, Wikipedia, and news domains. For summarization, we use datasets from news, literature, government, and science, including both single-document and multi-document summarization benchmarks. For information extraction, we use IEPile and group its constituent datasets into news, biomedical, science, and Wikipedia domains. Overall, the benchmark provides broad task and domain coverage, enabling a comprehensive evaluation of cross-domain model merging.

\subsection{LLM-as-a-judge Instruction}
\label{sec:prompt}

We use the following four prompt templates for LLM-as-a-judge evaluation using GPT4.1 for task of text summarization, lay paraphrasing, generative question answering and information extraction, respectively.

\begin{tcolorbox}[
  colback=gray!5,
  colframe=gray!40,
  boxrule=0.4pt,
  arc=1.5pt,
  left=5pt,
  right=5pt,
  top=4pt,
  bottom=4pt,
  fontupper=\footnotesize\ttfamily
]

\textbf{Text Summarization}
\textbf{System Prompt}

You are an impartial grader of \textbf{TEXT SUMMARIZATION} quality.

Given a \textbf{SOURCE} document, a \textbf{CANDIDATE} summary, and
(optionally) \textbf{REFERENCE} summaries, return a JSON object with
\texttt{score}, \texttt{verdict}, and \texttt{rationale}.

\textbf{Evaluation Dimensions} (priority order):
\begin{itemize}\setlength{\itemsep}{1pt}\setlength{\parskip}{0pt}
  \item \textbf{Faithfulness}: all claims supported by SOURCE.
  \item \textbf{Coverage}: salient points from SOURCE.
  \item \textbf{Coherence \& Fluency}: readable and grammatical.
  \item \textbf{Conciseness}: avoids unnecessary repetition.
\end{itemize}

\textbf{Scoring} ($[0,1]$):
\begin{itemize}\setlength{\itemsep}{1pt}\setlength{\parskip}{0pt}
  \item \textbf{1.0}: fully faithful; strong coverage of key points; coherent and concise.
  \item \textbf{0.8--0.9}: faithful with minor omissions/redundancy; generally coherent.
  \item \textbf{0.6--0.7}: mostly faithful with omissions OR some verbosity/awkwardness.
  \item \textbf{0.3--0.5}: partial coverage with minor coherence issues.
  \item \textbf{0.1--0.2}: weak coverage.
  \item \textbf{0.0}: largely hallucinated or irrelevant.
\end{itemize}

\textbf{Rules}:
\begin{itemize}\setlength{\itemsep}{1pt}\setlength{\parskip}{0pt}
  \item Judge \textbf{only} against SOURCE.
  \item Use REFERENCES \textbf{only} for salience.
  \item Do not penalize paraphrasing.
  \item Penalize unsupported claims.
  \item Verbatim repetition is allowed if faithful.
\end{itemize}

\textbf{Output} (\textbf{STRICT JSON ONLY}):
\begin{verbatim}
{"score":[0,1],
 "verdict":"Correct|Partial|Incorrect",
 "rationale":"<=25 words"}
\end{verbatim}

\textbf{User Prompt}

SOURCE DOCUMENT: \{question\}

REFERENCE SUMMARIES (optional): \{references\_block\}

CANDIDATE SUMMARY: \{candidate\}

\end{tcolorbox}

\begin{tcolorbox}[
  colback=gray!5,
  colframe=gray!40,
  boxrule=0.4pt,
  arc=1.5pt,
  left=5pt,
  right=5pt,
  top=4pt,
  bottom=4pt,
  fontupper=\footnotesize\ttfamily
]

\textbf{Lay Paraphrasing System Prompt}

You are an impartial grader of \textbf{PARAPHRASING} quality.

Given a \textbf{SOURCE} sentence, a \textbf{CANDIDATE} paraphrase, and
(optionally) \textbf{REFERENCE} paraphrases, return a JSON object with
\texttt{score}, \texttt{verdict}, and \texttt{rationale}.

\textbf{Evaluation Dimensions}:
\begin{itemize}\setlength{\itemsep}{1pt}\setlength{\parskip}{0pt}
  \item \textbf{Meaning Preservation}: semantic equivalence to SOURCE.
  \item \textbf{Fluency \& Grammaticality}: natural and correct language.
  \item \textbf{Paraphrase Diversity}: non-trivial rewording; penalize near-duplicates.
  \item \textbf{Factual Discipline}: no hallucinated or contradictory claims.
\end{itemize}

\textbf{Scoring} ($[0,1]$):
\begin{itemize}\setlength{\itemsep}{1pt}\setlength{\parskip}{0pt}
  \item \textbf{1.0}: faithful, fluent, clearly reworded.
  \item \textbf{0.8--0.9}: faithful and fluent; minor stylistic issues.
  \item \textbf{0.6--0.7}: mostly faithful but close wording or minor omissions.
  \item \textbf{0.3--0.5}: partially faithful.
  \item \textbf{0.1--0.2}: marginal; poor faithfulness or fluency.
  \item \textbf{0.0}: incorrect meaning or hallucinated content.
\end{itemize}

\textbf{Rules}:
\begin{itemize}\setlength{\itemsep}{1pt}\setlength{\parskip}{0pt}
  \item Judge \textbf{semantic equivalence}, not surface similarity.
  \item Ignore harmless stylistic variation.
  \item Penalize contradictions or added information.
\end{itemize}

\textbf{Output} (\textbf{STRICT JSON ONLY}):
\begin{verbatim}
{"score":[0,1],
 "verdict":"Correct|Partial|Incorrect",
 "rationale":"<=25 words"}
\end{verbatim}

\textbf{User Prompt}

SOURCE: \{source\}

REFERENCE PARAPHRASES (optional):
\{references\_block\}

CANDIDATE PARAPHRASE: \{candidate\}

\end{tcolorbox}

\begin{tcolorbox}[
  colback=gray!5,
  colframe=gray!40,
  boxrule=0.4pt,
  arc=1.5pt,
  left=5pt,
  right=5pt,
  top=4pt,
  bottom=4pt,
  fontupper=\footnotesize\ttfamily
]
\textbf{Question Answering}
\textbf{System Prompt}

You are an impartial grader of \textbf{QUESTION ANSWERING} quality.

Given a \textbf{QUESTION}, a \textbf{CANDIDATE} answer, and one or more
\textbf{REFERENCE} answers, return a JSON object with \texttt{score},
\texttt{verdict}, and \texttt{rationale}.

\textbf{Evaluation Criteria}:
\begin{itemize}\setlength{\itemsep}{1pt}\setlength{\parskip}{0pt}
  \item \textbf{Correctness}: semantically matches any reference answer.
  \item \textbf{Completeness}: includes required key details.
  \item \textbf{Consistency}: no contradictions or unsupported claims.
\end{itemize}

\textbf{Scoring} ($[0,1]$):
\begin{itemize}\setlength{\itemsep}{1pt}\setlength{\parskip}{0pt}
  \item \textbf{1.0}: fully correct with no contradictions.
  \item \textbf{0.8--0.9}: almost correct, very minor omissions or phrasing differences.
  \item \textbf{0.6--0.7}: mostly correct, contains the main idea but lacks details.
  \item \textbf{0.3--0.5}: somewhat relevant with minor errors.
  \item \textbf{0.1--0.2}: marginally related but mostly incorrect.
  \item \textbf{0.0}: completely incorrect or irrelevant.
\end{itemize}

\textbf{Rules}:
\begin{itemize}\setlength{\itemsep}{1pt}\setlength{\parskip}{0pt}
  \item Judge \textbf{semantic equivalence}, not exact wording.
  \item Ignore harmless extra details; penalize unsupported/contradictory claims.
  \item If references differ, accept any correct alternative consistent with common knowledge.
\end{itemize}

\textbf{Output} (\textbf{STRICT JSON ONLY}):
\begin{verbatim}
{"score":[0,1],
 "verdict":"Correct|Partial|Incorrect",
 "rationale":"<=25 words"}
\end{verbatim}

\textbf{User Prompt}

QUESTION: \{question\}

REFERENCE ANSWERS (one per line):
\{references\_block\}

CANDIDATE ANSWER: \{candidate\}

\end{tcolorbox}

\begin{tcolorbox}[
  colback=gray!5,
  colframe=gray!40,
  boxrule=0.4pt,
  arc=1.5pt,
  left=4pt,
  right=4pt,
  top=3pt,
  bottom=3pt,
  fontupper=\scriptsize\ttfamily
]

\textbf{Information Extraction System Prompt}

You are an impartial grader for \textbf{INFORMATION EXTRACTION} (event extraction).

\textbf{Inputs}:
\begin{itemize}\setlength{\itemsep}{0.5pt}
  \item \textbf{SOURCE} text.
  \item \textbf{CANDIDATE\_JSON}: possibly malformed JSON-like string
        \texttt{\{"event\_type":[\{"trigger":"...","arguments":\{...\}\},...]\}}.
  \item \textbf{REFERENCE\_JSONS}: one or more JSON-like strings with the same intent.
\end{itemize}

\textbf{Procedure}:
\begin{itemize}\setlength{\itemsep}{0.5pt}
  \item \textbf{Parse \& Repair}: recover best-effort JSON (accept single quotes, stray braces, repetitions).
  \item \textbf{Canonicalize}: lowercase; strip punctuation; collapse whitespace;
        lists de-duplicated (order irrelevant); \texttt{nan/n/a/""} $\rightarrow$ \texttt{NAN};
        \textbf{exact match only} after canonicalization.
  \item \textbf{Event Matching}: per event type, match events \textbf{1--to--1 by trigger};
        unmatched candidate $\rightarrow$ FP\_events; unmatched reference $\rightarrow$ FN\_events.
  \item \textbf{Argument Scoring} (matched events only):
        ref=\texttt{NAN} $\Rightarrow$ candidate value $\rightarrow$ FP\_args;
        ref string/list must be present $\Rightarrow$ TP\_args, else FN\_args;
        candidate extras $\rightarrow$ FP\_args.
\end{itemize}

If both reference and candidate contain no events, set score=1.0.

\textbf{Output} (\textbf{STRICT JSON ONLY}):
\begin{verbatim}
{"score":[0,1],"precision":[0,1],"recall":[0,1],"f1":[0,1],
 "verdict":"Correct|Partial|Incorrect",
 "diagnostics":{"tp_events":0,"fp_events":0,"fn_events":0,
                "tp_args":0,"fp_args":0,"fn_args":0},
 "rationale":"<=25 words"}
\end{verbatim}

\textbf{Rules}: no fuzzy matching; ignore harmless formatting; unrecoverable JSON $\Rightarrow$ score=0.

\textbf{User Prompt}

SOURCE: \{question\}

REFERENCE\_JSONS: \{references\_block\}

CANDIDATE\_JSON: \{candidate\}

\end{tcolorbox}

%\subsection{Parameter Settings} \label{sec:appendix_setting}

%\subsection{Domain-Level Ablation Results}
%\label{sec:appendix_ablation}

%We report the complete domain-level ablation results in this section. Unlike the main ablation table, which reports domain-averaged scores, the following tables show results for each individual domain and evaluation metric. The hierarchical architecture ablation further supports the distributional motivation of HWM: aggregating representations through task-level and global Wasserstein structure consistently outperforms a non-hierarchical global aggregation strategy, indicating that domain and task heterogeneity is better captured by hierarchical distributional modeling than by a flat aggregation scheme.

\begin{table}[p]
\centering
\caption{
Domain-level ablation results for the hierarchical architecture. BS denotes BERTScore F1, Valid. denotes prediction validity rate, MiF1 denotes Micro-F1, and LLM denotes LLM-as-a-judge. $\Delta$ denotes HWM minus the global architecture.
}
\label{tab:appendix_ablation_arch}
\scriptsize
\renewcommand{\arraystretch}{0.88}
\begin{tabular}{@{}lllrrr@{}}
\toprule
Task & Metric & Domain & Global & HWM & $\Delta$ \\
\midrule

\multirow{12}{*}{Summ.}
& \multirow{4}{*}{BERTScore}
  & Gov.  & 78.85 & 79.94 & +1.09 \\
& & Lit.  & 74.83 & 76.53 & +1.70 \\
& & News & 75.65 & 77.30 & +1.65 \\
& & Sci.  & 70.58 & 77.75 & +7.17 \\
\cmidrule(lr){2-6}
& \multirow{4}{*}{SARI}
  & Gov.  & 34.39 & 41.15 & +6.76 \\
& & Lit.  & 39.86 & 39.54 & -0.32 \\
& & News & 37.86 & 42.09 & +4.23 \\
& & Sci.  & 36.00 & 39.61 & +3.61 \\
\cmidrule(lr){2-6}
& \multirow{4}{*}{LLM}
  & Gov.  & 57.97 & 67.28 & +9.31 \\
& & Lit.  & 50.38 & 53.32 & +2.94 \\
& & News & 54.60 & 60.79 & +6.19 \\
& & Sci.  & 66.30 & 78.40 & +12.10 \\

\midrule

\multirow{12}{*}{Lay}
& \multirow{4}{*}{BERTScore}
  & Bio.  & 79.67 & 81.74 & +2.07 \\
& & News & 81.59 & 79.27 & -2.32 \\
& & Sci.  & 80.27 & 80.65 & +0.38 \\
& & Wiki & 82.01 & 83.33 & +1.32 \\
\cmidrule(lr){2-6}
& \multirow{4}{*}{SARI}
  & Bio.  & 40.38 & 40.19 & -0.19 \\
& & News & 33.62 & 38.92 & +5.30 \\
& & Sci.  & 32.39 & 34.52 & +2.13 \\
& & Wiki & 31.45 & 36.87 & +5.42 \\
\cmidrule(lr){2-6}
& \multirow{4}{*}{LLM}
  & Bio.  & 65.52 & 70.14 & +4.62 \\
& & News & 39.43 & 43.49 & +4.06 \\
& & Sci.  & 43.65 & 43.71 & +0.06 \\
& & Wiki & 38.42 & 44.77 & +6.35 \\

\midrule

\multirow{12}{*}{QA}
& \multirow{4}{*}{BERTScore}
  & Bio.  & 74.02 & 77.82 & +3.80 \\
& & News & 76.53 & 77.42 & +0.89 \\
& & Web  & 76.96 & 77.80 & +0.84 \\
& & Wiki & 77.73 & 78.30 & +0.57 \\
\cmidrule(lr){2-6}
& \multirow{4}{*}{METEOR}
  & Bio.  & 21.18 & 21.23 & +0.05 \\
& & News & 22.60 & 28.21 & +5.61 \\
& & Web  & 23.74 & 28.45 & +4.71 \\
& & Wiki & 28.92 & 27.11 & -1.81 \\
\cmidrule(lr){2-6}
& \multirow{4}{*}{LLM}
  & Bio.  & 66.92 & 67.17 & +0.25 \\
& & News & 42.57 & 45.40 & +2.83 \\
& & Web  & 61.09 & 62.62 & +1.53 \\
& & Wiki & 71.88 & 79.69 & +7.81 \\

\midrule

\multirow{12}{*}{IE}
& \multirow{4}{*}{Valid.}
  & Bio.  & 67.96 & 85.71 & +17.75 \\
& & News & 85.35 & 98.25 & +12.90 \\
& & Sci.  & 72.06 & 94.10 & +22.04 \\
& & Wiki & 80.13 & 94.12 & +13.99 \\
\cmidrule(lr){2-6}
& \multirow{4}{*}{MiF1}
  & Bio.  & 6.16  & 12.34 & +6.18 \\
& & News & 28.67 & 36.60 & +7.93 \\
& & Sci.  & 5.75  & 15.27 & +9.52 \\
& & Wiki & 22.92 & 37.14 & +14.22 \\
\cmidrule(lr){2-6}
& \multirow{4}{*}{LLM}
  & Bio.  & 13.53 & 22.90 & +9.37 \\
& & News & 29.50 & 44.87 & +15.37 \\
& & Sci.  & 11.88 & 20.48 & +8.60 \\
& & Wiki & 34.50 & 69.27 & +34.77 \\

\bottomrule
\end{tabular}
\end{table}

\begin{table*}[p]
\centering
\caption{
Domain-level ablation results for the number of cluster centroids. BS denotes BERTScore F1, Valid. denotes prediction validity rate, MiF1 denotes Micro-F1, and LLM denotes LLM-as-a-judge.
}
\label{tab:appendix_ablation_centroids}
\scriptsize
\renewcommand{\arraystretch}{0.86}
\begin{tabular}{@{}lllrrrrrr@{}}
\toprule
Task & Metric & Domain
& \multicolumn{3}{c}{HWM}
& \multicolumn{3}{c}{HWM$^\ast$} \\
\cmidrule(lr){4-6}
\cmidrule(lr){7-9}
& & 
& $K=128$ & $K=256$ & $K=512$
& $K=128$ & $K=256$ & $K=512$ \\
\midrule

\multirow{12}{*}{Summ.}
& \multirow{4}{*}{BS}
  & Gov.  & 77.00 & 77.84 & 79.94 & 75.83 & 75.83 & 77.28 \\
& & Lit.  & 74.15 & 74.85 & 76.53 & 73.42 & 73.40 & 77.41 \\
& & News & 75.26 & 77.07 & 77.30 & 72.74 & 76.95 & 77.37 \\
& & Sci.  & 74.72 & 76.54 & 77.75 & 70.58 & 76.07 & 79.48 \\
\cmidrule(lr){2-9}
& \multirow{4}{*}{SARI}
  & Gov.  & 38.11 & 38.20 & 41.15 & 34.33 & 34.67 & 39.48 \\
& & Lit.  & 37.79 & 38.18 & 39.54 & 36.43 & 36.48 & 39.96 \\
& & News & 40.35 & 42.96 & 42.09 & 39.50 & 42.75 & 42.24 \\
& & Sci.  & 37.73 & 39.36 & 39.61 & 36.00 & 39.27 & 40.21 \\
\cmidrule(lr){2-9}
& \multirow{4}{*}{LLM}
  & Gov.  & 56.87 & 67.58 & 67.28 & 56.49 & 57.08 & 63.00 \\
& & Lit.  & 30.84 & 36.65 & 53.32 & 50.59 & 50.59 & 53.23 \\
& & News & 54.85 & 54.77 & 60.79 & 59.70 & 61.73 & 66.13 \\
& & Sci.  & 62.53 & 69.32 & 78.40 & 75.94 & 78.07 & 80.90 \\

\midrule

\multirow{12}{*}{Lay}
& \multirow{4}{*}{BS}
  & Bio.  & 80.27 & 80.18 & 81.74 & 82.27 & 82.58 & 83.48 \\
& & News & 77.39 & 79.58 & 79.27 & 79.24 & 79.52 & 79.84 \\
& & Sci.  & 80.52 & 80.32 & 80.65 & 79.24 & 79.56 & 79.12 \\
& & Wiki & 83.20 & 83.46 & 83.33 & 74.07 & 74.50 & 77.70 \\
\cmidrule(lr){2-9}
& \multirow{4}{*}{SARI}
  & Bio.  & 40.07 & 40.22 & 40.19 & 42.80 & 42.97 & 42.48 \\
& & News & 33.60 & 33.43 & 38.92 & 40.30 & 40.82 & 40.97 \\
& & Sci.  & 30.63 & 30.89 & 34.52 & 30.30 & 31.67 & 32.39 \\
& & Wiki & 31.68 & 31.14 & 36.87 & 28.00 & 30.46 & 33.93 \\
\cmidrule(lr){2-9}
& \multirow{4}{*}{LLM}
  & Bio.  & 70.24 & 70.35 & 70.14 & 62.80 & 63.10 & 64.37 \\
& & News & 41.82 & 42.88 & 43.49 & 40.00 & 40.60 & 45.06 \\
& & Sci.  & 42.67 & 42.42 & 43.71 & 36.50 & 38.40 & 36.49 \\
& & Wiki & 36.80 & 38.19 & 44.77 & 34.10 & 39.90 & 46.48 \\

\midrule

\multirow{12}{*}{QA}
& \multirow{4}{*}{BS}
  & Bio.  & 72.77 & 72.62 & 77.82 & 74.14 & 74.20 & 77.64 \\
& & News & 77.32 & 77.36 & 77.42 & 73.19 & 76.60 & 80.39 \\
& & Web  & 76.49 & 77.10 & 77.80 & 80.27 & 80.15 & 79.54 \\
& & Wiki & 77.47 & 78.42 & 78.30 & 79.30 & 79.98 & 78.00 \\
\cmidrule(lr){2-9}
& \multirow{4}{*}{METEOR}
  & Bio.  & 20.77 & 21.02 & 21.23 & 74.96 & 75.86 & 19.93 \\
& & News & 21.92 & 21.46 & 28.21 & 22.21 & 20.57 & 31.50 \\
& & Web  & 27.05 & 27.55 & 28.45 & 27.50 & 27.46 & 30.91 \\
& & Wiki & 26.99 & 26.99 & 27.11 & 24.90 & 27.56 & 27.65 \\
\cmidrule(lr){2-9}
& \multirow{4}{*}{LLM}
  & Bio.  & 65.42 & 65.83 & 67.17 & 60.30 & 63.47 & 65.13 \\
& & News & 42.66 & 41.89 & 45.40 & 45.50 & 46.73 & 49.51 \\
& & Web  & 56.96 & 52.07 & 62.62 & 58.02 & 66.53 & 67.19 \\
& & Wiki & 76.84 & 79.74 & 79.69 & 76.67 & 79.74 & 81.18 \\

\midrule

\multirow{12}{*}{IE}
& \multirow{4}{*}{Valid.}
  & Bio.  & 82.73 & 80.66 & 85.71 & 82.03 & 84.00 & 85.66 \\
& & News & 85.45 & 97.40 & 98.25 & 97.55 & 98.75 & 98.90 \\
& & Sci.  & 93.97 & 94.11 & 94.10 & 90.35 & 90.13 & 92.20 \\
& & Wiki & 94.84 & 96.19 & 94.12 & 91.98 & 97.10 & 95.56 \\
\cmidrule(lr){2-9}
& \multirow{4}{*}{MiF1}
  & Bio.  & 10.57 & 7.09  & 12.34 & 10.09 & 10.24 & 11.81 \\
& & News & 26.73 & 32.07 & 36.60 & 34.02 & 39.59 & 41.32 \\
& & Sci.  & 13.77 & 9.39  & 15.27 & 13.06 & 14.20 & 15.38 \\
& & Wiki & 28.71 & 31.49 & 37.14 & 36.97 & 39.21 & 39.66 \\
\cmidrule(lr){2-9}
& \multirow{4}{*}{LLM}
  & Bio.  & 12.24 & 18.83 & 22.90 & 24.26 & 23.36 & 23.65 \\
& & News & 40.91 & 42.10 & 44.87 & 44.98 & 44.80 & 47.22 \\
& & Sci.  & 14.18 & 18.18 & 20.48 & 20.57 & 20.38 & 22.98 \\
& & Wiki & 69.16 & 64.36 & 69.27 & 68.69 & 63.19 & 70.39 \\

\bottomrule
\end{tabular}
\end{table*}

\begin{table*}[p]
\centering
\caption{
Domain-level ablation results for probing-set size. BS denotes BERTScore F1, Valid. denotes prediction validity rate, MiF1 denotes Micro-F1, and LLM denotes LLM-as-a-judge. $\Delta$ denotes the full probing setup minus the corresponding small probing setup.
}
\label{tab:appendix_ablation_probing}
\scriptsize
\renewcommand{\arraystretch}{0.86}
\begin{tabular}{@{}lllrrrrrr@{}}
\toprule
Task & Metric & Domain
& \multicolumn{3}{c}{HWM}
& \multicolumn{3}{c}{HWM$^\ast$} \\
\cmidrule(lr){4-6}
\cmidrule(lr){7-9}
& &
& Small & Full & $\Delta$
& Small & Full & $\Delta$ \\
\midrule

\multirow{12}{*}{Summ.}
& \multirow{4}{*}{BS}
  & Gov.  & 77.87 & 79.94 & +2.07 & 75.80 & 77.28 & +1.48 \\
& & Lit.  & 73.36 & 76.53 & +3.17 & 73.28 & 77.41 & +4.13 \\
& & News & 74.61 & 77.30 & +2.69 & 77.02 & 77.37 & +0.35 \\
& & Sci.  & 74.66 & 77.75 & +3.09 & 77.40 & 79.48 & +2.08 \\
\cmidrule(lr){2-9}
& \multirow{4}{*}{SARI}
  & Gov.  & 38.45 & 41.15 & +2.70 & 34.99 & 39.48 & +4.49 \\
& & Lit.  & 37.73 & 39.54 & +1.81 & 36.48 & 39.96 & +3.48 \\
& & News & 39.93 & 42.09 & +2.16 & 42.99 & 42.24 & -0.75 \\
& & Sci.  & 37.50 & 39.61 & +2.11 & 39.38 & 40.21 & +0.83 \\
\cmidrule(lr){2-9}
& \multirow{4}{*}{LLM}
  & Gov.  & 62.28 & 67.28 & +5.00 & 65.10 & 63.00 & -2.10 \\
& & Lit.  & 44.11 & 53.32 & +9.21 & 49.95 & 53.23 & +3.28 \\
& & News & 58.42 & 60.79 & +2.37 & 65.30 & 66.13 & +0.83 \\
& & Sci.  & 77.12 & 78.40 & +1.28 & 78.71 & 80.90 & +2.19 \\

\midrule

\multirow{12}{*}{Lay}
& \multirow{4}{*}{BS}
  & Bio.  & 79.52 & 81.74 & +2.22 & 83.15 & 83.48 & +0.33 \\
& & News & 77.55 & 79.27 & +1.72 & 80.57 & 79.84 & -0.73 \\
& & Sci.  & 80.08 & 80.65 & +0.57 & 73.40 & 79.12 & +5.72 \\
& & Wiki & 81.25 & 83.33 & +2.08 & 81.88 & 77.70 & -4.18 \\
\cmidrule(lr){2-9}
& \multirow{4}{*}{SARI}
  & Bio.  & 38.98 & 40.19 & +1.21 & 44.01 & 42.48 & -1.53 \\
& & News & 33.87 & 38.92 & +5.05 & 41.82 & 40.97 & -0.85 \\
& & Sci.  & 30.59 & 34.52 & +3.93 & 25.63 & 32.39 & +6.76 \\
& & Wiki & 33.75 & 36.87 & +3.12 & 29.34 & 33.93 & +4.59 \\
\cmidrule(lr){2-9}
& \multirow{4}{*}{LLM}
  & Bio.  & 66.43 & 70.14 & +3.71 & 58.81 & 64.37 & +5.56 \\
& & News & 42.75 & 43.49 & +0.74 & 43.71 & 45.06 & +1.35 \\
& & Sci.  & 43.01 & 43.71 & +0.70 & 36.19 & 36.49 & +0.30 \\
& & Wiki & 42.52 & 44.77 & +2.25 & 44.95 & 46.48 & +1.53 \\

\midrule

\multirow{12}{*}{QA}
& \multirow{4}{*}{BS}
  & Bio.  & 74.25 & 77.82 & +3.57 & 74.32 & 77.64 & +3.32 \\
& & News & 78.21 & 77.42 & -0.79 & 80.52 & 80.39 & -0.13 \\
& & Web  & 71.31 & 77.80 & +6.49 & 80.11 & 79.54 & -0.57 \\
& & Wiki & 77.80 & 78.30 & +0.50 & 78.91 & 78.00 & -0.91 \\
\cmidrule(lr){2-9}
& \multirow{4}{*}{METEOR}
  & Bio.  & 22.25 & 21.23 & -1.02 & 11.03 & 19.93 & +8.90 \\
& & News & 30.04 & 28.21 & -1.83 & 31.36 & 31.50 & +0.14 \\
& & Web  & 21.14 & 28.45 & +7.31 & 27.16 & 30.91 & +3.75 \\
& & Wiki & 23.03 & 27.11 & +4.08 & 25.99 & 27.65 & +1.66 \\
\cmidrule(lr){2-9}
& \multirow{4}{*}{LLM}
  & Bio.  & 59.80 & 67.17 & +7.37 & 63.76 & 65.13 & +1.37 \\
& & News & 40.40 & 45.40 & +5.00 & 46.34 & 49.51 & +3.17 \\
& & Web  & 56.04 & 62.62 & +6.58 & 68.32 & 67.19 & -1.13 \\
& & Wiki & 70.20 & 79.69 & +9.49 & 76.73 & 81.18 & +4.45 \\

\midrule

\multirow{12}{*}{IE}
& \multirow{4}{*}{Valid.}
  & Bio.  & 81.10 & 85.71 & +4.61 & 84.57 & 85.66 & +1.09 \\
& & News & 94.15 & 98.25 & +4.10 & 92.65 & 98.90 & +6.25 \\
& & Sci.  & 89.29 & 94.10 & +4.81 & 90.13 & 92.20 & +2.07 \\
& & Wiki & 84.53 & 94.12 & +9.59 & 92.85 & 95.56 & +2.71 \\
\cmidrule(lr){2-9}
& \multirow{4}{*}{MiF1}
  & Bio.  & 8.25  & 12.34 & +4.09 & 10.86 & 11.81 & +0.95 \\
& & News & 32.10 & 36.60 & +4.50 & 38.88 & 41.32 & +2.44 \\
& & Sci.  & 9.91  & 15.27 & +5.36 & 15.00 & 15.38 & +0.38 \\
& & Wiki & 33.60 & 37.14 & +3.54 & 38.68 & 39.66 & +0.98 \\
\cmidrule(lr){2-9}
& \multirow{4}{*}{LLM}
  & Bio.  & 23.86 & 22.90 & -0.96 & 24.52 & 23.65 & -0.87 \\
& & News & 38.88 & 44.87 & +5.99 & 46.95 & 47.22 & +0.27 \\
& & Sci.  & 13.37 & 20.48 & +7.11 & 18.61 & 22.98 & +4.37 \\
& & Wiki & 63.63 & 69.27 & +5.64 & 68.32 & 70.39 & +2.07 \\

\bottomrule
\end{tabular}
\end{table*}
\end{document}